# Understanding the Feedforward Artificial Neural Network Model From the Perspective of Network Flow

Dawei Dai[1], Weimin Tan[1], Hong Zhan[2] {dw_dai@163.com}
1. School of Computer Science, Shanghai Key Laboratory of Data Science, Fudan University, Shanghai, China.
2. School of Automation Science and Engineering, South China University of Technology, GuangZhou, China

## Abstract:

In recent years, deep learning based on artificial neural network (ANN) has achieved great success in pattern recognition. However, there is no clear understanding of such neural computational models [1], for instance, for a trained ANN-classifier, we have no idea of why some classes are easy to be predicted correctly, but some are difficult to be predicted correctly. In this paper, we try to unravel "black-box" structure of classifiers and explain the mechanism of classification decisions made by classifiers from network flow. Specifically, we consider the feed forward artificial neural network as a network flow model, which consists of many directional class-pathways. Each class-pathway encodes one class. The class-pathway of a class is obtained by connecting the activated neural nodes in each layer from input to output, where activation value of neural node (node-value) is defined by the weights of each layer in a trained ANN-classifier. From the perspective of the class-pathway, training an ANN-classifier can be regarded as the formulation process of class-pathways of different classes. By analyzing the the distances of each two class-pathways in a trained ANN-classifiers, we try to answer the questions, why the classifier performs so? The smaller the distance of the class-pathways between two classes is, the higher the probability of the predicted error each other for these two classes will be. Furthermore, we can use the analysis as a new way to measure the performance of classifier and compare them despite their high prediction accuracy reporting on the small test sets. At last, from the neural encodes view, we define the importance of each neural node through the class-pathways, which is helpful to optimize the structure of a classifier. Experiments for two types of ANN model including multi-layer perceptron (MLP) and convolutional neural network (CNN) verify that the network flow based on class-pathway is a reasonable explanation for ANN models.
*Keywords*: Ann, cnn, mlp, deep learning, network flow;

# \section{Introduction}

Deep neural networks (DNNs), especially convolutional neural networks (CNN), have been widely used in recent years and demonstrated excellent performance in pattern recognition fields, such as speech recognition, image classification, and face recognition. The researches indicate that DNNs can achieve outstanding performance on much more challenging classification tasks, for instance, Deep CNN models have shown record beating performance on NORB, CIFAR-10, ImageNet and PASCAL VOC datasets [2, 3, 4]. Most of the researches focus on the application-level in different fields and also some of them try to explore what factors are responsible for the improvement in performance, such as neuron modeling [5] for avoiding the gradient fade, designing effective loss functions for different tasks, reasonable network structure for improving the network structure from psychologyor brain, regularization strategies [6] for avoiding over-fitting and so on. These works have made the DNNs a great success in pattern recognition fields.

Despite this encouraging progress, there is still little insight into the internal operation and behavior of DNNs, or how they achieve such good performance [1, 7, 8, 11, 12]. We have to try different kinds of tricks in every specific task to achieve good performance. We can say that the neural network model is still a "black-box" structure. As a result, it is necessary for us to understand its internal working mechanism. According to the basic ideas of current approaches for understanding DNNs, we can divide them into two categories: one is to visualize what does the network learn by finding neurons with maximally activation-intensity for an input image [1, 7, 8, 9]; another is to visualize how sensitive the classification accuracy is to a specific region in the input image in order to explain a particular classification made by the network [1, 10, 11]. These two categories can effectively reveal whether the network is trained well (check the feature maps in each layer) or which part of the input image is sensitive to the network. However, these approaches do not explain why a well-trained network model achieves good performance in some data sets, while badly in other data sets? For instance, for a trained multi-classifier (class-1, class-2, ... , class-i,..., class-n), why some classes are hard to be predicted correctly (having high prediction error rate) and why some classes are easily predicted (having low error rate)? When the samples from the class-i are predicted to other class labels incorrectly, which the class labels will easily be? Why will be easy to predict incorrectly to several specific classes for one specific class? From a scientific standpoint, this is deeply unsatisfactory. Without clear understanding of how and why they work well or bad, the development of better models is reduced to trial-and-error.

Based on the observation of brain energy expenditure, a research [13] assumes that the encoding and decoding of neurons are sparse and scattered. Neuroscience [14] concludes that only 1-4% of neurons are activated at the most time. Based on the theories and discoveries in neuroscience, rectified linear unit (RELU) model [15] better approximates biological nerve activation function. Typically, about 25%-30% neurons are activated at once forward calculation in ANN employing RELU neural model, i.e. Only a part of neurons is activated for once forward calculation, which is like a "path" in the network through the activated neurons and their connection weights.

In this paper, we aim to interpret the above mechanism of the forward neural network from the perspective of network flow, which consists of lots of directional class-pathways. A class-pathway can be thought that the input information is propagated layer by layer along a "path" from input nodes (neurons in input layer) to a specific output node (one neuron in output layer). Here, we consider each neuron in network as a node. The neurons in input layer are regarded as start nodes, and the neurons in output layer are output nodes. The output nodes represent the class labels for ANN-classifiers. From this perspective, training an ANN-classifier can be regarded as the formulation process of class-pathways for different classes. Through analysis the distance of two class-pathways for all classes in ANN-classifiers, we try to answer the above questions.

## \section{Related works}

According to the essential ideas of current algorithms, we broadly classify them into two categories: *learned feature visualization* and *class saliency visualization*. The former focuses on the visualizing features that learned by the network; the latter aims at highlights important areas in a given input image for a certain class.

**Learned feature visualization:** Research efforts have worked on visualizing what the network learns by finding neurons with maximally activation-intensity for an input image [7, 8, 9]. This category has been reported for example in [7], which measures how sensitive the classification accuracy is to small variations in pixel values of an input image. Research [10] for visualizing intermediate feature layers based on Deconvolutional Network (deconvnet) technique. [16] for reducing the dimensions of SIFT. [1] for compressing the image-level descriptors. Such approaches are able to remove the redundant information, requiring only a few bits for an interest point. However, these methods require feature extraction and vector quantization, which is with heavy computational cost for mobile devices with limited memory and battery. Besides, sending the compressed local descriptors to the cloud-end is able to achieve a low bit-rate query delivery, but it also loses global features such as colors, textures, object shapes, and the spatial relationship among local features.

**Class saliency visualization:** One such instance-specific method is class saliency visualization proposed in [7], who measure how sensitive the classification score is to small changes in pixel values, by computing the partial derivative of the class score with respect to the input features using standard back propagation. They also show that there is a close connection to using deconvolutional networks for visualization, proposed by [1]. Other methods include [9], who compare the activation of a unit when a specific input is fed forward through the net to reference activation for that unit. Studies [10, 11] also generate interesting visualization results for individual inputs, but are both not as closely related to our method as the two papers mentioned above. Another analysis [1] make: they estimate the importance of input pixels by visualizing the probability of the (correct) class as a function of a gray patch occluding parts of the image.

In the field of medical image classification specifically, a widely used method for visualizing feature importance is to simply plot the weights of a linear classifier [17, 18], or the p-values of these weights (determined by permutation testing) [19]. These are independent of the input image, as argued in [20], interpreting these weights can be misleading in general.

# \section{Approach}

We could imagine that each neuron was analogized to a water container, connections between neurons were analogized to the pipelines, and values of connection weights were just like different size of pipelines. The larger the size of pipeline is, the larger the amount of water through the pipeline is. Positive value of connection weights is analogized to that water flowing into the container; Negative value of weights is analogized to that the water flowing out of the container. For a container, if the amount of water out of the container is more than into the container, no water of this container flow into downstream containers.

Once forward calculation process of ANN-classifier just like water (information) flowing into or out of the containers from top to down through the pipelines. We can modify the size of pipeline layer by layer to guide water flow direction, which form the different *pathways* of water flow from the top containers to down containers. From this perspective, training an ANN-classifier can be regarded as the formulation process of a specific *pathway* for each class (*class-pathway*). Each *class-pathway* (information flowing from container in input layer to a specific container in output layer) corresponds to a class, and class-pathway is represented by the specific activated neural nodes in each layer and their connections. As shown in Fig.1(a) and (b), two *class-pathways* for two classes were formed after training.

Here, we used directed graph to describe the above process. We take each neuron as a neural node (water container) and take neural connections (pipelines) as unidirectional pathway, then neural network can be analogy of a directed graph including start nodes, hide nodes and output nodes, which propagate the information from the input nodes to the output nodes. We consider the feed forward artificial neural network as a network flow model, which consists of many directional *class-pathways*.

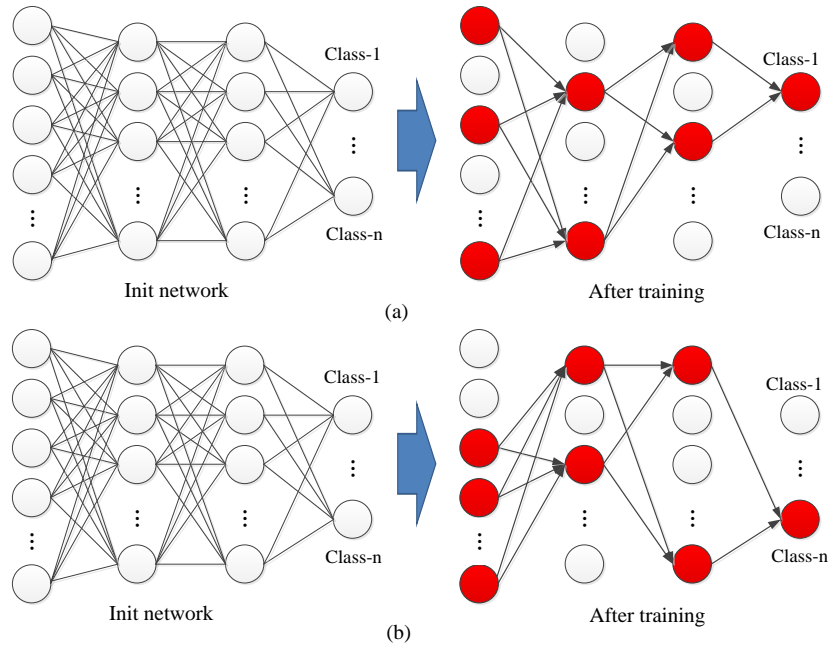

Figure 1. *Class-pathways* of a classifier. (a) After training, forming the *class-pathway* for Class-1;(b) After training, forming the *class-pathway* for Class-n.

## \subsection{Rectified Linear Units}

The Rectified Linear Units (Relu), an approximate neural model, comes from the sparseness research on the working human brain neurons. Research indicates that 95% - 99% of human brain neurons are typically idle. Less working neurons mean smaller computational complexity, which are less likely to be over-fitted. For the Relu neuron model, when the value of input is negative, the model output is 0, which corresponds to the inhibitory state of the biological neurons and has no effect on the postsynaptic neurons. When the input value is positive means the neuron activated. The larger the value is, the greater the activation intensity will be, which corresponds to the lager firing rate of biological neurons [15]. In a Relu based neural network model, a little part of neurons is activated in the process of each forward calculation. It has a certain biological basis whether from the perspective of a single neuron or the composed neural network. Furthermore, the Relu based neural network shows good performance in a variety of applications, which has become one of the mainstream neuron models in artificial neural network [21]. Therefore, the analysis of this paper is based on the Relu based forward neural network.

\subsection{Define the Node-value of network flow for MLP model}

In a well-trained classifier, each neuron has learned a specific feature, which can be obtained by the linear combination of the features that all neurons learned in the upper layer. Recent works have shown that the lower level features will be acquired by the neurons, which are more closely to the input layer, and the higher level features can be learned by the neurons, which are more closely to the output layer. The features of the samples from the same one class have a great degree of similarity on a measure (the basic assumption of machine learning). Therefore, when the different samples from the same one class are calculated by a trained neural network, the distribution of activated neural nodes and non-activated neurons in the network will be similar. That is to say, in a well-trained network model, samples from the same one class should take the similar *class-pathway* through the network in high probability and eventually converge to an output neural node that represents this class label (which is the maximum output neuron).

Fig.3 shows the directed graph model for a trained MLP model, in which the red nodes have positive activation value, and 0 otherwise. A class-pathway consists of the red nodes. Those nodes with 0 activation nodes that have no contributions to the nodes in the following layer, which are not included in the *class-pathway*. We use $VL_k[j]$ to denote the normalized activation value of the $j^{th}$ node in the $k^{th}$ layer. Thus $VL_k[j]$ denotes the probability of an input sample belonging to $j^{th}$ class. Given an input sample from $j^{th}$ class, the ideal case is that $VL_k[j]=1$, and others (except $j^{th}$ node) are 0.

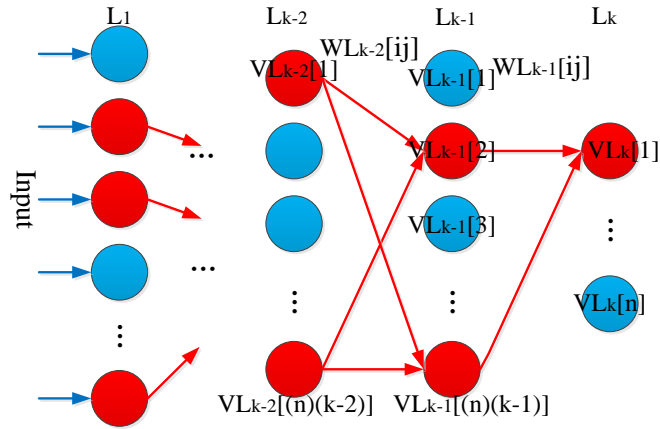

Figure 2 Network flow model for MLP

Here, we formerly formulate the network flow model for MLP. Equation (1) demonstrates the feed-forward computation of MLP. The $VL_k[j]$ achieves maximum value when the weight vector has the same direction as the output activation of nodes in the $L_{k-1}$ layer, i.e., Eq (3) holds. If $VL_{k-1}[i] < 0$, then $VL_{k-1}[i]=0$, which is the output characteristic of Relu model. $VL_{k-1}[i] = 0$ denotes that the $i^{th}$ node in $(k-1)^{th}$ layer makes no contribution to the nodes in next layer. Similarly, from Eq (2), we can infer that in order to make the intensity value of the first node in the $L_{k-1}$ layer to be the largest, the intensity value of the $L_{k-2}$ layer nodes should be the weight vector: $(WL_{k-2}[1i], WL_{k-2}[2i],..., WL_{k-2}[(k-2)i])$. However, $VL_{k-1}[i]$ is not always the maximum, which can be regarded as the multiplier coefficient, we obtain the intensity value of the $L_{k-2}$ layer node: $(WL_{-2}[1i], WL_{k-2}[2i],..., WL_{k-2}[(k-2)i])*VL_{k-1}[i]$. Since, each active node in the $L_{k-1}$ layer corresponds to a set of nodes-value in L$_{k-2}$ layer, the final strength value of the $L_{k-2}$ layer node is obtained as shown in Eq (4). The node-value in $L_1, L_2, ..., L_{k-3}$ nodes can be obtained in the same way.

$$VL_k[j] = F(\sum_{i=1}^{i=n} VL_{k-1}[i] * WL_{k-1}[ij]) \quad (1)$$

$$VL_{k-1}[j] = \sum_{i=1}^{i=n_{k-2}} VL_{k-2}[i] * WL_{k-2}[ij] \quad (2)$$

$$VL_{k-1}[i] = WL_{k-1}[ij], i = 1, 2, ..., n_{k-1} \quad (3)$$

$$VL_{k-2}[i] = \sum_{j=1}^{j=n_{k-2}} VL_{k-1}[j] * WL_{k-2}[ij] \quad (4)$$

In this way, for a trained MLP network model, we can infer all node-value for class-j. In idea case, the input sample from class-j passes through the selected nodes with positive node-value in the neural network and end at the j[th] node in output layer.

\subsection{Define the Node-value of network flow for CNN model}

In this section, we will explicitly show the network flow model for CNN. Fig.3 shows a widely used structure of CNN, which consists of convolutional layers, pooling layers, and fully connected layers for the layer-3, layer-4, and layer-5, which can be seen as a MLP model. Thus, we can get the nodes-value according to eq (1-4). The layer-3 is the expanded version of the layer-2. Thus, in the feature map sequences of pooling layer-2, each n4*n4 feature map corresponds to consecutive n4 * n4 neurons in the layer-3.

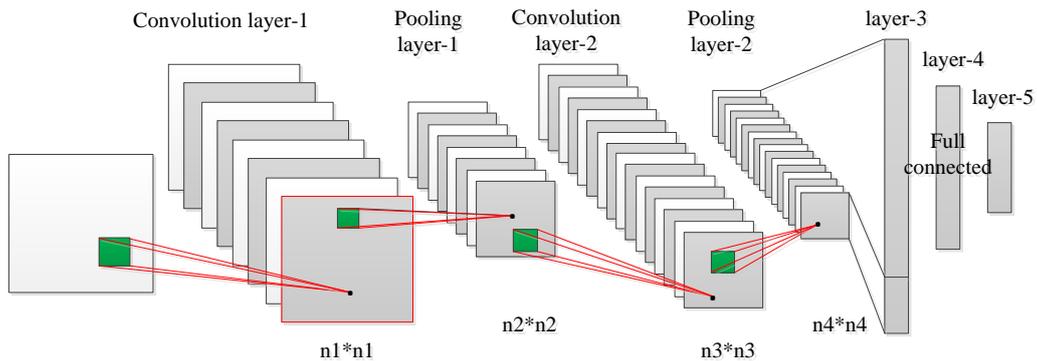

Figure 3 A classical structure of CNN Network

The convolutional structure is shown in Fig.4 (a). Each feature map is considered as a neural node, as shown in Fig.4(c). We still use node-value to represent the intensify value of each node. We obtain each node-value in pooling layer-2 by averaging the intensity values of all the corresponding nodes in layer-3, as shown in Eq.(5), where $V_{layer3}[i]$ is the i[th] node-value in the layer-3. Since, each pooling layer is obtained by down-sampling the corresponding convolutional layer. In general, the pooling layer still keeps the features of the corresponding convolutional layer unchanged. As a result, the intensity value of each feature map (node-value) in the convolutional layer-2 can be obtained by the node-value in pooling layer-2 multiplied a constant coefficient as in Eq.(6), where k within the range 0~1.

As shown in Fig.4 (b), each convolution kernel is regarded as a weight corresponding to MLP. We use the average of value in one convolution kernel as a connection weight. Afterwards, Fig.4 (a) can be simplified to a standard MLP structure, as shown in Fig.4(d). Therefore, we can employ the calculating way mentioned in previous section to compute all the nodes-value (intensity value of each feature map) in the upper pooling layer. Based on Eq.(7), we can calculate the node-value of each class for a trained CNN network model.

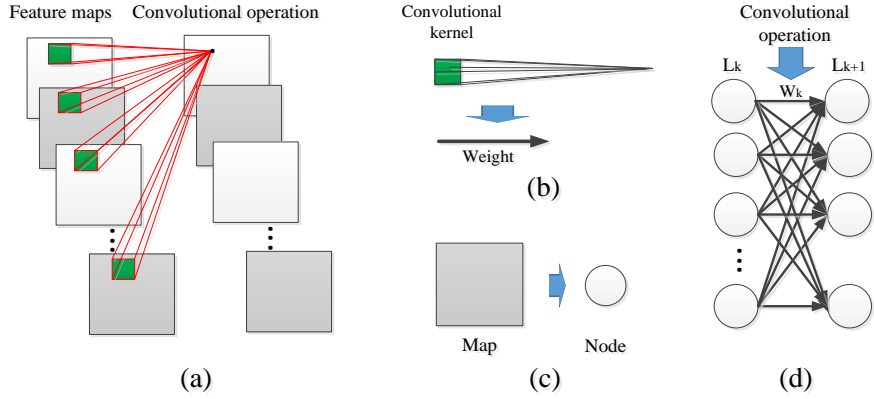

Figure 4 Simplification of convolutional networks

$$V_{Player\ 2}[i] = \frac{1}{n4*n4}\sum_{i=(i-1)*n4*n4}^{i=i*n4*n4} V_{layer\ 3}[i] \quad (5)$$

$$V_{Clayer\ 2}[i] = k * V_{Player\ 2}[i] \quad (6)$$

$$V_{Clayer\ 1}[i] = \sum_{j=1}^{j=n} V_{Clayer\ 2}[j] * W_1[ij] \quad (7)$$

## \subsection{Define the Class-pathway and analysis}

In the human brain, neurons encode different information by different firing rate. The different node-value of Relu model can be considered as the different firing rate of neuron. A trained n-classifier model typically has *n* output neurons. Ideally, the input sample from class-i makes the activation value of the $i^{th}$ neuron in output layer to 1 and others to 0, which the n-classifier model determines that the current input sample is from the $i^{th}$ class. The previous two sections have demonstrated that we can obtain a set of nodes-value for each class, i.e., *Vci:* {...,$V_{L1}[i]$,..., $V_{L2}[j]$,...,$V_{Li}[k]$,...}, where $V_{Li}[k]$ denotes the intensity value of the $k^{th}$ node in the $i^{th}$ layer. The information only goes through the nodes that with positive node-value. Here, we define the set *Vci class-pathway* for class-i. In this way, we can obtain *n* class-pathways in a trained n-classifier, and each *class-pathway* encodes one class.

We define the distance between the $i^{th}$ and the $j^{th}$ class based on their *class-pathways*, where $\|Dij\|_2$ is defined as Eq.(8). $\|Dij\|_2$ can be used to quantitative represent the distance or similarity of the $i^{th}$ class and the j[th] class. Intuitively, the smaller the distance $\|Dij\|_2$ between the i[th] class and j[th] class is (where $\|Dij\|_2 < \|Dik\|_2$, j!=k), the larger the error rate of the i[th] class predicted as the j[th] class is. If the distances $\|Dik\|_2$ between the i[th] class and other classes (where k=1, 2, ..., n) is obviously smaller than the distances $\|Djk\|_2$ (where k=1,2,...,n), the trained model will perform badly for the samples from the i[th] class than from the j[th] class.

$$\|Dij\|_2 = \sqrt{\sum_{index=0}^{node\_num} |(Vi[index] - Vj[index])|^2} \quad (8)$$

# \section{Experiments and Analysis}

We illustrate the proposed approach by a trained MLP-classifier model and a CNN-classifier model on the MINIST dataset. For the MLP classifier, we employ a structure of 784*600*600*10 with two hidden layers, where the neuron models in the hidden layer and the output layer are Relu and Softmax respectively. For the CNN classifier, it consists of two convolutional layers, a fully connected hide layer, and a softmax output layer. The feature maps in the first and second convolutional layers are 20 and 80, respectively, and the hidden layer consists of 400 neurons. Note that both these two classifiers achieve over 97% prediction accuracy on the test set of MINIST.

## \subsection{Analysis for MLP model}

For MLP model, we can obtain the class-pathway *Vci* (i=0, 1, ..., 9) of each class based on Eq.(1-4). Fig.6 shows the node-values in the two hidden layers. The node-value of the same node for different classes are different. The nodes with values close to zero can be regard as non-active. From Fig.5 (a) and (b), we observe that the non-active nodes account for a certain proportion, i.e., the information of the previous layer does not flow through those non-active nodes. Actually, it may be in this way that the MLP model encodes different classes.

Table 1 demonstrates the distance matrix *D*, where a cell *Dij* in D denotes the Euclidean distance between the class-pathways of the i[th] and j[th] classes. Table 2 indicates the average distance between class-i and other classes. To obtain more test examples, we add random noises to MINIST dataset (including training and testing sets, validation sets) and obtain 70,000 test examples. Our trained MLP model tests on this enlarged test set, where there are 1954 examples incorrectly predicted. Table 3 shows the number of examples of class-i (column) incorrectly predicted as class-j (i, j=0, 1, 2, ..., 9). Table 1 and Table 3 statistically demonstrate that the number of incorrectly predicted examples to the 4 classes that class-pathway closest to the class i accounts for about 80.76% of all the incorrectly predicted examples; the number of examples to the 5 classes closest to the class i accounts for about 91.61% of all the incorrectly predicted examples. Table 2 also shows that class-3, class-8, and class-9 have the minimal average distance with other class-pathways. The number of examples incorrectly predicted as class-3, class-8, and class-9 accounts for 80.8% of all the incorrectly predicted examples for all classes.

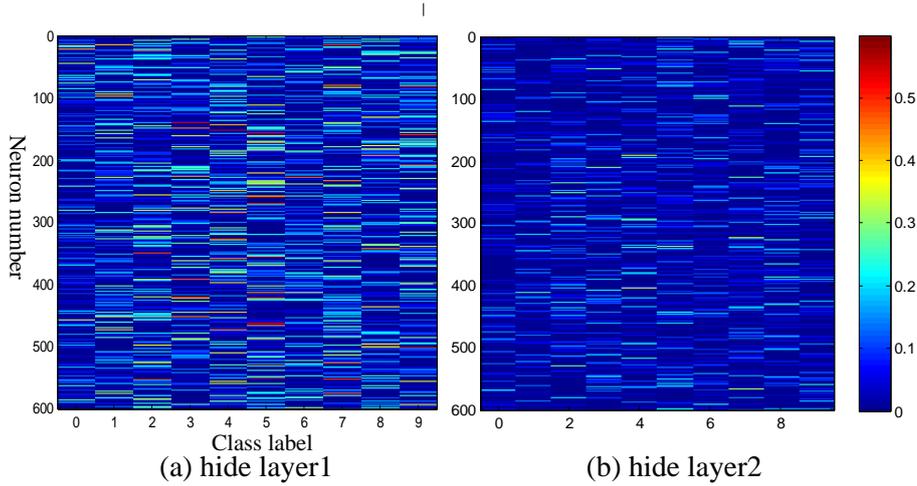

(a) hide layer1      (b) hide layer2

Figure 5. class-pathways for each class in MLP, x-axis indicates class label and y-axis indicates node number. (a)(b) Node-value of each class in the first and second hide layers respectively.

Table1 Distance matrix D of *class-pathway*

|  | Class-0 | Class-1 | Class-2 | Class-3 | Class-4 | Class-5 | Class-6 | Class-7 | Class-8 | Class-9 |
|---|---|---|---|---|---|---|---|---|---|---|
| Class-0 | 0 | 5.7164 | 6.9109 | 6.0568 | 6.7163 | 6.3376 | 6.3940 | 6.7947 | 6.3784 | 5.3476 |
| Class-1 | 5.7164 | 0 | 6.8504 | 5.2794 | 6.7224 | 6.1653 | 6.5219 | 6.2917 | 6.0810 | 5.2730 |
| Class-2 | 6.9109 | 6.8505 | 0 | 6.1134 | 7.0224 | 6.2142 | 5.9786 | 5.8219 | 5.0135 | 5.7142 |
| Class-3 | 6.0568 | 5.2794 | 6.1134 | 0 | 5.9708 | 5.3145 | 6.5511 | 5.9542 | 4.6559 | 4.3453 |
| Class-4 | 6.7163 | 6.7225 | 7.0224 | 5.9708 | 0 | 7.0202 | 6.4134 | 7.7140 | 6.4586 | 6.4732 |
| Class-5 | 6.3376 | 6.1653 | 6.2142 | 5.3145 | 7.0202 | 0 | 7.0934 | 6.6108 | 4.9500 | 5.9626 |
| Class-6 | 6.3940 | 6.5219 | 5.9786 | 6.5511 | 6.4134 | 7.0934 | 0 | 6.1975 | 6.1283 | 5.8657 |
| Class-7 | 6.7948 | 6.2917 | 5.8219 | 5.9542 | 7.7140 | 6.6108 | 6.1975 | 0 | 5.8266 | 5.5541 |
| Class-8 | 6.3784 | 6.0810 | 5.0135 | 4.6559 | 6.4586 | 4.9500 | 6.1283 | 5.8266 | 0 | 5.5541 |
| Class-9 | 5.3477 | 5.2730 | 5.7142 | 4.3453 | 6.4742 | 5.9626 | 5.8657 | 5.1876 | 5.5541 | 0 |

Table2 Average distance of any one class with other classes

| Class-0 | Class-1 | Class-2 | Class-3 | Class-4 | Class-5 | Class-6 | Class-7 | Class-8 | Class-9 |
|---|---|---|---|---|---|---|---|---|---|
| **6.2948** | 6.1002 | 6.1822 | **5.5824** | **6.7235** | 6.1854 | **6.3493** | 6.2655 | **5.6718** | **5.5248** |

Table3. The number of examples from any one class incorrectly predicted to other classes

|  | Class-0 | Class-1 | Class-2 | **Class-3** | Class-4 | Class-5 | Class-6 | Class-7 | **Class-8** | **Class-9** |
|---|---|---|---|---|---|---|---|---|---|---|
| Class-0 | 0 | 0 | 7 | 1 | 1 | 2 | 1 | 0 | 24 | 6 |
| Class-1 | 0 | 0 | 14 | 36 | 1 | 0 | 5 | 2 | 504 | 1 |
| Class-2 | 8 | 2 | 0 | 28 | 5 | 1 | 3 | 1 | 42 | 3 |
| Class-3 | 0 | 0 | 7 | 0 | 0 | 15 | 0 | 1 | 45 | 4 |
| Class-4 | 2 | 1 | 8 | 8 | 0 |  | 8 | 1 | 81 | 64 |
| Class-5 | 6 | 0 | 2 | 54 | 1 | 0 | 13 | 1 | 63 | 8 |
| Class-6 | 12 | 1 | 5 | 2 | 6 | 12 | 0 | 0 | 95 | 0 |
| Class-7 | 10 | 4 | 97 | 228 | 4 | 10 | 0 | 0 | 69 | 50 |
| Class-8 | 8 | 0 | 3 | 16 | 0 | 5 | 2 | 2 | 0 | 1 |
| Class-9 | 9 | 2 | 1 | 43 | 5 | 11 | 6 | 6 | 142 | 0 |

## \subsection{Analysis for CNN}

For CNN model, we can obtain the set *Vci* (i=0,1,...,9) of each class according to Eq(6-8). Fig.6 shows the node-values in the convolutional layer and fully layers. The nodes with values close to zero can be regard as non-active. The active nodes form a *class-pathway* for each class. Actually, we may say that the CNN model encodes different classes by such way.

Table4 shows the distance matrix *D*, an element *Dij* in *D* denotes the euclidean distance between the class-pathways of the i$^{th}$ and j$^{th}$ classes. Table 5 indicates the average distance between class-i with other classes. We get an enlarged set (70000 samples) through adding random noises to MINIST dataset. The CNN model tests on this enlarged test set, where there are 2909 examples incorrectly predicted. Table5 shows the number of examples of class-i (column) incorrectly predicted as class-j (i, j = 0, 1, 2, ... , 9). Table4 and Table6 statistically demonstrate that the number of incorrectly predicted examples to the 4 classes that their *class-pathways* closest to the class-i accounts for about 81.23% of all the incorrectly predicted examples; the number of examples to the 5 classes closest to the class-i accounts for about 86.52% of all the incorrectly predicted examples. Table5 also shows that class-3, class-4, class-8, and class-9 have the minimal average distance with other *class-pathways*. The number of examples incorrectly predicted as class-3, class-4, class-8, and class-9 accounts for 89.69% of all the incorrectly predicted examples for all classes.

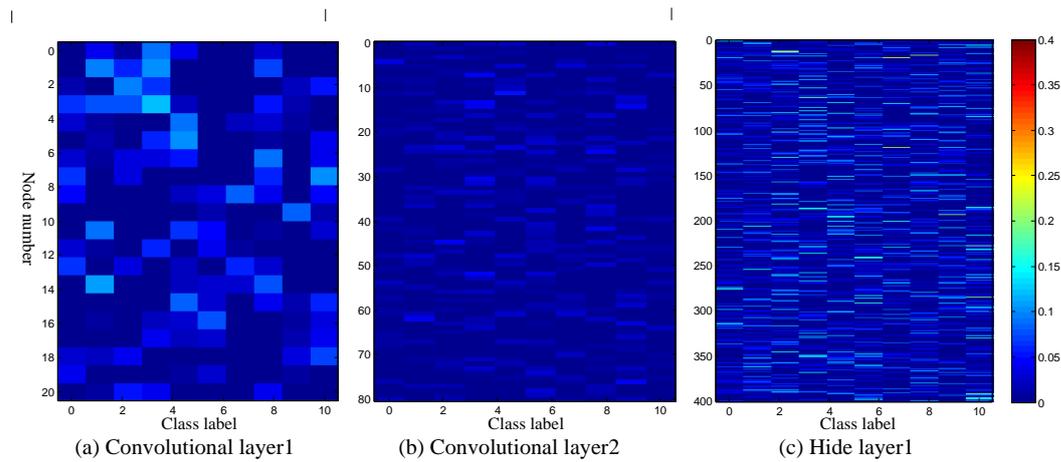

Figure 6. *class-pathway*s for each class in CNN model, x-axis indicates class label and y-axis indicates node number. (a)(b) Node-value of different class-pathway in the first and second convolutional layer respectively; (c) Node-value of different class-pathway in hide units of fully connected layer.

Table4 Distance matrix D of *class-pathway*

|         | Class-0 | Class-1 | Class-2 | Class-3 | Class-4 | Class-5 | Class-6 | Class-7 | Class-8 | Class-9 |
|---------|---------|---------|---------|---------|---------|---------|---------|---------|---------|---------|
| Class-0 | 0       | 0.9551  | 1.0255  | 1.0179  | 0.9494  | 0.9638  | 1.0203  | 1.0848  | 1.0167  | 0.9363  |
| Class-1 | 0.9551  | 0       | 0.9649  | 0.8384  | 0.9002  | 0.9586  | 0.9714  | 0.9638  | 0.9489  | 0.8592  |
| Class-2 | 1.0255  | 0.9649  | 0       | 0.9644  | 0.9445  | 0.9443  | 0.9769  | 0.9476  | 0.8102  | 0.8642  |
| Class-3 | 1.0179  | 0.8384  | 0.9644  | 0       | 0.8477  | 0.8824  | 0.9939  | 0.9812  | 0.9005  | 0.7316  |
| Class-4 | 0.9494  | 0.9002  | 0.9445  | 0.8447  | 0       | 0.9817  | 0.9585  | 1.0227  | 0.9115  | 0.8830  |
| Class-5 | 0.9638  | 0.9586  | 0.9443  | 0.8824  | 0.9817  | 0       | 1.0360  | 1.0211  | 0.8365  | 0.9147  |
| Class-6 | 1.0203  | 0.9714  | 0.9769  | 0.9939  | 0.9585  | 1.0360  | 0       | 1.0445  | 0.9576  | 0.9534  |
| Class-7 | 1.0848  | 0.9489  | 0.8102  | 0.9005  | 0.9115  | 0.8365  | 0.9576  | 0       | 0.9141  | 0.8629  |
| Class-8 | 1.0167  | 0.9489  | 0.8102  | 0.9005  | 0.9115  | 0.8365  | 0.9576  | 0.9141  | 0       | 0.8229  |

| Class-9 | 0.9363 | 0.8592 | 0.8642 | 0.7316 | 0.8830 | 0.9147 | 0.9534 | 0.8629 | 0.8229 | 0 |

Table5 Average distance of any one class with other classes

| Class-0 | Class-1 | Class-2 | Class-3 | Class-4 | Class-5 | Class-6 | Class-7 | Class-8 | Class-9 |
| --- | --- | --- | --- | --- | --- | --- | --- | --- | --- |
| 8.9698 | 8.3604 | 8.4424 | 8.1579 | 8.3393 | 8.5392 | 8.9135 | 8.8437 | 8.1188 | 7.8282 |

Table6 The number of examples from any one class incorrectly predicted to other classes

|  | Class-0 | Class-1 | Class-2 | Class-3 | Class-4 | Class-5 | Class-6 | Class-7 | Class-8 | Class-9 |
| --- | --- | --- | --- | --- | --- | --- | --- | --- | --- | --- |
| Class-0 | 0 | 0 | 2 | 2 | 3 | 2 | 32 | 1 | 36 | 3 |
| Class-1 | 1 | 0 | 20 | 10 | 7 | 0 | 7 | 5 | 1391 | 0 |
| Class-2 | 4 | 0 | 0 | 13 | 16 | 0 | 7 | 3 | 84 | 0 |
| Class-3 | 2 | 0 | 6 | 0 | 1 | 14 | 0 | 1 | 76 | 3 |
| Class-4 | 0 | 0 | 0 | 0 | 0 | 0 | 5 | 1 | 27 | 3 |
| Class-5 | 4 | 0 | 2 | 22 | 5 | 0 | 36 | 2 | 110 | 1 |
| Class-6 | 2 | 1 | 0 | 0 | 12 | 5 | 0 | 0 | 29 | 0 |
| Class-7 | 7 | 5 | 50 | 38 | 24 | 3 | 0 | 0 | 134 | 8 |
| Class-8 | 1 | 1 | 2 | 0 | 0 | 2 | 2 | 0 | 0 | 1 |
| Class-9 | 10 | 2 | 0 | 48 | 150 | 28 | 5 | 17 | 352 | 0 |

## \subsection{Optimization of neural network structure from neural encode view}

Each class-pathway corresponds to a class, which can be seen as some kind of neural encode. And each class-pathway contains a set of node-value. Node-value can be considered as a threshold, which controls how much information can be went through. Here, neural encode of a specific class is achieved by a set of node-value (*class-pathway*). If some node-value in a specific class-pathway equals or close to zero, which mean that these nodes have no or little contribution to the nodes in the next layers, in other words, these nodes can be removed.

We can optimize the structure of a classifier from the perspective of neural information encode. In an ANN-classifier, if the node-value of node is zero or little in every class-pathway, then we may say that this node is not needed and can be removed. For a 10-classifier (MLP and CNN model), according to eq.9, $V_{ci}$ is the node-value set of class-i, Vector *Vneu* {...,*Vneu_Lij*, ...} can be considered as the importance measure of each node in the ANN-classifier, Vneu_Lij indicates the importance of the $j^{th}$ neuron in the $i^{th}$ layer, the smaller of the *Vneu_Lij* is, the less important of this node is. We remove the less important nodes to optimize the ANN-classifier but not affect its performance.

$$V_{neu} = \sum_{i=1}^{i=10} Vi \quad (9)$$

For the above MLP classifier (structure: 784*600*600) with the 2.03% error rate on the test dataset, first, we get vector *Vneu* for the model and sort the elements in *Vneu* from small to large. In Fig7(a), we gradually increase the cutting number of nodes that from node with the smallest value in *Vneu* in the first and second hide layers, and the error rate increases; Fig7(b) indicates that error rate of this classifier will not have a great change with cutting number in the first hide layer less than 50 and less than 200 in the second layer. For the above CNN classifier (structure: 784*20*80*400*10) with the 1.07% error rate on the test dataset, Fig7 (c) and (d) also indicate that error rate of the classifier will not have a great change with cutting number in the second convolutional layer less than 10 and less than 150 in the hide layer.

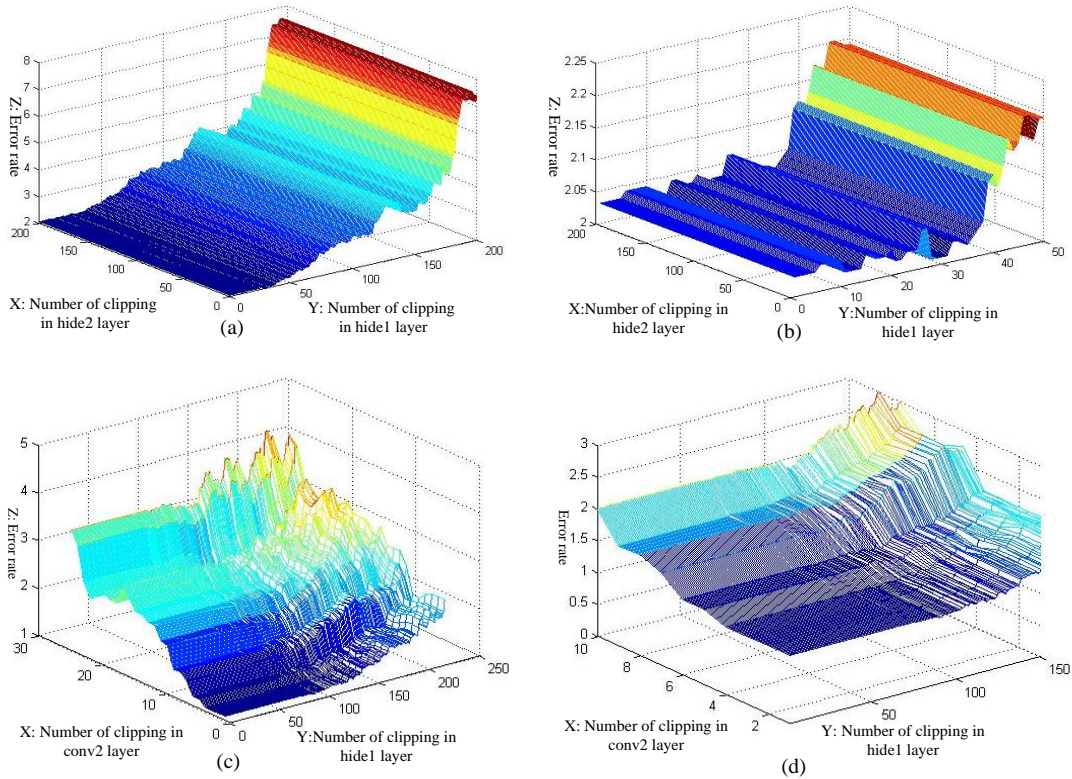

Figure 7. Optimize the structure of MLP and CNN classifier. (a) Error rate on the test dataset with cutting the increasing number of nodes according to their importance in the first and second hide layers of MLP classifier; (b) Subset of (a), the cutting number within a certain range, error rate change little; (c) and (d) for CNN classifier.

## \subsection{Experiments Analysis}

From the analysis of the class-pathways for MLP and CNN, we can define 10 class-pathways, which is made up of nodes-value, and each class-pathway in ANN-classifier encode one class; if the samples from class-i are predicted incorrectly, it is likely the several specific classes having small class-pathway distance with class-i. If most of the distance of class-i with other classes are smaller than other classes', it is possible that this ANN-classifier will has much bad performance for class-i than other classes.

We consider class-pathway as a kind of neural encode of each class in ANN-classifier. The node-value is analogized to the firing rate of biological neuron. If node-value of a specific node is zero, which indicates that this neuron is non-active and not take part in encoding information. The larger the node-value is, the larger the firing rate is. The biological neurons encode information by different firing rate. We define the importance measure of each node by class-pathway, and remove the less important nodes. The experiment proves its rationality.

## \section{Discussion}

We summarize our work. First of all, this paper made a vivid metaphor, when the samples go through the neural network layer by layer for calculation, which is just like to inject the different amounts of water into the top nodes, and then flow through layer by layer to output nodes by pipeline (connection weights). If the amount of water of the output node-i is much more than other output nodes, then we could say the sample belong to class-i. In the paper, we describe such process as a network flow model. From this perspective, the essence of the learning process of a n-ANN-classifier is to form $n$ class-pathways from input nodes to output nodes. In the paper, we define the class-pathways and the distance of two class-pathways. If the samples from class-i were predicted incorrectly (i != j), we can say that it is easy to be predicted as one of the classes having smaller distances. Since, the smaller the distances of two class-pathways, the more similar the two class-pathway and the two classes are. For a well-trained classifier, why the correct rate for the samples from a specific class is lower than other classes? It is possible that the distances of this class with other classes are smaller than other classes'.

In the resent years, deep learning models based on the neural network has achieved great success in many fields and is constantly entering into new fields. However, there is no clear understanding of internal working mechanism of neural network based model, for instance, for a well-trained ANN-classifier, we have no idea of which classes are easy to predict? Which classes are difficult to predict?, and why? In order to achieve good performance, we have to attempt different kinds of tricks for every specific task. As a result, we can say that the neural network model is still a "black-box" structure. Our works in the paper are focusing on this issue, we attempt to interpret the internal mechanism of ANN-classifier. We analogy the neural network structure to a network flow model, and then define the class-pathway for every class. Each class-pathway including the node with different activation can be seen as a kind of neural encode. Through analyzing the distance of class-pathways, we obtain the difference and similarity of these encodes in the ANN-classifier for one class with other classes. We attempt to interpret and understand what happens inter the ANN-classifier from network flow or neural encode perspective.

**Refs**